\documentclass{article}

\usepackage[T1]{fontenc}
\usepackage{microtype}
\usepackage{amsmath,amssymb}
\usepackage{hyperref}
\usepackage{graphicx}
\usepackage{booktabs}
\usepackage[preprint]{icml2027}

\hypersetup{pdfsubject={Evaluation protocol}}
\newcommand{\doi}[1]{\href{https://doi.org/#1}{doi:\,\nolinkurl{#1}}}
\newcommand{\authoremail}[1]{{\footnotesize\href{mailto:#1}{\nolinkurl{#1}}}}
\icmltitlerunning{MedFork: Active Multimodal Evidence Verification}
\icmlkeywords{Interactive medical AI, evaluation protocols, evidence verification}

\begin{document}

\twocolumn[
  \icmltitle{MedFork: A Protocol for Evaluating Active Multimodal\texorpdfstring{\\}{ }%
    Verification in Clinical Consultations}
  \begin{icmlauthorlist}
    \begin{tabular}{@{}c@{\hspace{1.75em}}c@{\hspace{1.75em}}c@{}}
      \icmlauthor{Zhihao Tang}{tsinghua,jdhealth} &
      \icmlauthor{Yunsong Liu}{shandong} &
      \icmlauthor{Chao Li}{jdhealth} \\
      \authoremail{tzh@mail.tsinghua.edu.cn} &
      \authoremail{liuys5@alumni.sysu.edu.cn} &
      \authoremail{lichao37@jd.com} \\
      \addlinespace
      \icmlauthor{Xuyang Cao}{jdhealth} &
      \icmlauthor{Jintao Fei}{jdhealth} &
      \icmlauthor{Jun Zhao}{jdhealth} \\
      \authoremail{newxuyangcao@gmail.com} &
      \authoremail{fei-jintao@outlook.com} &
      \authoremail{zhaojun10@jd.com}
    \end{tabular}
  \end{icmlauthorlist}
  \icmlaffiliation{tsinghua}{Tsinghua University.\enspace}
  \icmlaffiliation{jdhealth}{JD Health International Inc.\enspace}
  \icmlaffiliation{shandong}{Department of Radiation Oncology, Shandong Cancer Hospital and Institute,
    Shandong First Medical University, Shandong Academy of Medical Sciences, Jinan, Shandong 250117, P.~R.~China}
  \icmlcorrespondingauthor{Zhihao Tang}{\mbox{\href{mailto:tzh@mail.tsinghua.edu.cn}{tzh@mail.tsinghua.edu.cn}}}
  \vskip 0.20in
]

\printAffiliationsAndNotice{}

% Defer Figure 1 to the top of the first page's right column.
\suppressfloats[t]
\begin{figure}[t]
  \centering
  \includegraphics[width=\columnwidth]{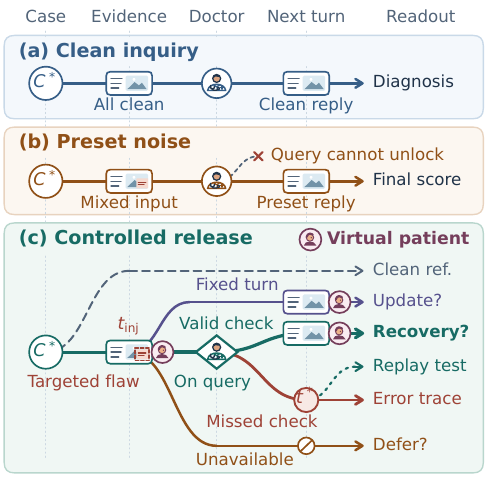}
  \caption{\textbf{Paired evidence release from one case.} (a) Evidence acquisition in a clean inquiry. (b) Example of a prescribed noisy follow-up. (c) The flawed branches share the same targeted flaw at $t_{\mathrm{inj}}$. Verified evidence is delivered on a fixed turn ($L_0$), contingent on a question meeting a general or specific threshold ($L_1$ or $L_2$), or kept unavailable ($L_\infty$). The clean reference supplies the original evidence at flaw entry. Replay from before the first observable error turn $t^*$ compares a targeted query with an unmodified continuation.}
  \label{fig:worldlines}
\end{figure}

\begin{abstract}
In a multimodal clinical consultation, a model may need to identify and verify an unreliable history item or report field before reaching a diagnosis. Evaluation with only a final diagnosis can miss whether the model requested the evidence that resolved the flaw. MedFork specifies an evaluation protocol that branches consultations from the same dialogue and simulator state. A controlled flaw is introduced into a known field, and verified evidence is delivered on a fixed schedule, released after a qualifying question, or kept unavailable. The protocol records recognition, verification requests, actual evidence release, and diagnostic responses. Paired replay compares a targeted verification query with an unmodified continuation from the same pre-error state. A hand-worked case illustrates how to apply the rules and interpret their outputs. The contribution is the protocol and its scoring definitions. Empirical validation of the simulator and model behavior remains open.
\end{abstract}

\section{Introduction}

Interactive medical AI uses multimodal large language models (MLLMs) to support clinical diagnosis through multi-turn consultations. These consultations rely on multimodal evidence, combining verbal patient narratives with visual artifacts (e.g., laboratory reports and clinical photographs)~\citep{amie}. Static medical question answering presents evidence in an initial prompt~\citep{medqa}, while a consultation can require gathering verbal facts, inspecting visual report fields, and updating a diagnosis across turns. An evaluation of this process should account for whether a model requests verification when the evidence it receives is unreliable.

Interactive medical benchmarks measure how MLLMs obtain evidence across turns~\citep{agentclinic,ddxtrace,hx2dx,threemdb,amie}. AgentClinic-NEJM, for example, includes a condition in which the model requests an image during the consultation~\citep{agentclinic}. Other benchmarks study responses to flawed evidence, including image corruption, misleading context, and text-image conflict~\citep{medvigil,mccxr,trickortreat,rimbench}. Multi-turn evaluations also study imperfect patient communication~\citep{meddialbench,verisim}, scripted sequences with evidence stressors~\citep{medbenchv5}, fabricated evidence presented in a later turn~\citep{forgery}, and degraded clinical photos~\citep{imcbench}.

MedFork addresses a specific question within this literature. Given a known flaw in a verbal fact or visual report field, does the doctor MLLM identify the issue, ask a question that can resolve it, and revise or defer its conclusion as evidence becomes available? Patient misreporting motivates the verbal setting~\citep{pamis}. A blurred or mismatched report field provides a visual counterpart. A model may revise its diagnosis when a clear report arrives automatically and still fail to request the report in the first place. MedFork compares these behaviors using different release rules for the same targeted field.

A straightforward approach to building this evaluation paradigm is to feed all clean evidence, clinical images, and a scenario prompt into an MLLM. The prompt specifies what flaw to simulate and when to release verified evidence, letting the MLLM role-play a virtual patient freely. To compare different conditions (e.g., clean versus flawed evidence), each condition is run as a separate consultation from turn one. Once each consultation finishes, an evaluator checks the final diagnosis or reads the finished dialogue to locate where the doctor MLLM erred.

This approach raises two design problems. \textbf{(1) Control of the patient state.} A role-play prompt containing the full record gives the generator access to fields that the protocol intends to keep hidden. Free-form replies can also change background details. \textbf{(2) Comparison across dialogue branches.} Consultations started independently can diverge before a flaw appears. A final diagnosis and a single finished trace also leave open what would happen if a targeted question were asked at an earlier turn~\citep{clintraj,replayrca}.

This paper specifies \textbf{MedFork}, a snapshot-and-fork evaluation protocol with a \emph{vault-isolated patient agent} and a \emph{fork-based turn-level auditor}. The patient agent keeps withheld facts and report fields outside the generator's input and applies controlled flaws to the target fields. A state machine implements fixed-turn, question-triggered, and unavailable release rules. The auditor branches consultations from the same pre-flaw state at $t_{\mathrm{inj}}$, logs the model's behavior across turns, and marks the first observable error turn $t^*$ when one can be identified. The protocol defines repeated rollouts for policy comparisons and paired replay for estimating recovery under a specified intervention.

The protocol makes three contributions.
\begin{itemize}
\setlength{\itemsep}{2pt}
\item \textbf{Task formulation.} Active multimodal verification is defined as a conditional evidence-acquisition task over a flawed verbal fact or visual report field in a multi-turn consultation.
\item \textbf{Controlled release.} A vault-isolated patient specification defines field-level perturbations, a clean reference, and four release policies covering fixed-turn, general-query, specific-query, and unavailable evidence.
\item \textbf{Turn-level audit.} Shared-prefix branches and paired replay define how to record recognition, requests, release, diagnostic responses, and recovery under a targeted query. Section~\ref{sec:worked_example} applies the rules to a hand-worked case.
\end{itemize}
\section{Problem Setup}
\label{sec:problem-setup}

\noindent\textbf{Definition 2.1 (Clean Base Cases and Flawed Evidence).}
Consider a set of clean base cases $\mathcal{C} = \{C_1^*, \dots, C_N^*\}$. Each base case $C^* = (\mathcal{E}^*, Y^*)$ pairs a ground-truth clinical diagnosis $Y^*$ with $M$ clean evidence items $\mathcal{E}^* = \{e_1^*, \dots, e_M^*\}$. Each item $e_k^* = (v_k^*, m_k^*)$ consists of a verbal fact $v_k^*$, a visual report field $m_k^* = (I_k^*, b_k)$ locating field $b_k$ in a clinical image or report $I_k^*$, or both. Both $v_k$ and $m_k$ can take four states in $\mathcal{A} = \{\text{clean}, \text{missing}, \text{degraded}, \text{false}\}$: accurate evidence, omitted facts or cropped fields, vague recall or blurry fields, and confident falsehoods or altered report fields. An evaluation scenario injects flaws into a target subset $\mathcal{F} \subseteq \mathcal{E}^*$ (a single-field flaw when $|\mathcal{F}| = 1$, or combined flaws when $|\mathcal{F}| > 1$) while keeping $\mathcal{E}^* \setminus \mathcal{F}$ clean; cross-modal conflicts and self-contradictions arise when incompatible states appear for the same item.

\noindent\textbf{Definition 2.2 (Multi-Turn Consultation and Prefix-Locked Forking).}
In an interactive multimodal consultation, a doctor MLLM $D$ communicates with a patient agent $P$ across $T$ turns before issuing a final diagnostic response. At turn $t$, $D$ produces $u_t$ given $H_{<t}$, and $P$ replies with $r_t$ (text with zero or more visual artifacts), forming $H_{\le t} = ((u_1, r_1), \dots, (u_t, r_t))$. The controller updates the evidence state $\mathcal{S}_t$ from the preceding state, $u_t$, and the case's presentation schedule and release rule. A rollout policy $\pi$ fixes the patient rule and uses the same doctor model, with an explicit override at one turn for the intervention in Section~\ref{sec:3.2.3}. A full trajectory $\tau = (H_{\le T}, \hat{Y})$ includes the final response $\hat{Y}$ and follows the factorized interaction distribution:

\begin{equation}
\begin{aligned}
&P_\pi(\tau \mid C^*, \mathcal{F})\\
&\quad = \Pr\nolimits_D(\hat{Y} \mid H_{\le T})
  \prod_{t=1}^{T} \bigl[\Pr\nolimits_D(u_t \mid H_{<t}; \pi)\\
&\qquad {}\times \Pr\nolimits_P(r_t \mid H_{<t}, u_t, \mathcal{S}_t; \pi)\bigr].
\end{aligned}
\label{eq:1}
\end{equation}

Capturing the dialogue state before turn $t$ as a snapshot $\Sigma_{t-1} = (H_{<t}, \mathcal{S}_{t-1})$, the forking operator $\mathrm{Fork}(\Sigma_{t-1}, \pi, K)$ locks the identical prefix $H_{<t}$ and draws $K$ independent continuations under the same configuration:

\begin{equation}
\begin{aligned}
\tau^{(j)} &= H_{<t} \oplus \tau^{(j)}_{t:T},\\
\tau^{(j)}_{t:T} &\stackrel{\text{i.i.d.}}{\sim}
  P_\pi(\tau_{t:T} \mid \Sigma_{t-1}), \quad j \in \{1,\dots,K\}.
\end{aligned}
\label{eq:2}
\end{equation}

\noindent\textbf{Definition 2.3 (Active Multimodal Verification).}
Given a clean base case $C^*$ and target flaw set $\mathcal{F}$, $t_{\mathrm{inj}}$ is the first scheduled presentation of any targeted item, where the policies can first differ. From the shared snapshot $\Sigma_{t_{\mathrm{inj}}-1}$, paired worldlines are forked across a clean reference $\pi_{\text{clean}}$ and four release policies $\rho \in \{L_0, L_1, L_2, L_\infty\}$. The protocol defines the expected outcome under each policy from the identical prefix:

\begin{equation}
\begin{aligned}
\mathcal{J}(\pi_\rho \mid \Sigma_{t_{\mathrm{inj}}-1})
&= \mathbb{E}_{\tau \sim P_{\pi_\rho}(\cdot \mid \Sigma_{t_{\mathrm{inj}}-1})}
  \!\left[\Phi_\rho(\tau, Y^*)\right],\\
&\quad \rho \in \{\text{clean}, L_0, L_1, L_2, L_\infty\},
\end{aligned}
\label{eq:3}
\end{equation}

For $\rho \neq L_\infty$, $\Phi_\rho(\tau, Y^*)$ is 1 if the final response matches the reference diagnosis under a prespecified scoring rubric and 0 otherwise. Under $L_\infty$, it is 1 if the final response identifies the unresolved evidentiary gap and withholds a firm conclusion, and 0 otherwise. The policies distinguish proactive release ($L_0$), release in reply to a qualifying inquiry ($L_1$ or $L_2$), and unavailable verification ($L_\infty$). The last policy has a distinct endpoint, deferral under unavailable evidence.

\section{Methodology}
\label{sec:3}

\begin{figure*}[!t]
  \centering
  \includegraphics[width=\textwidth]{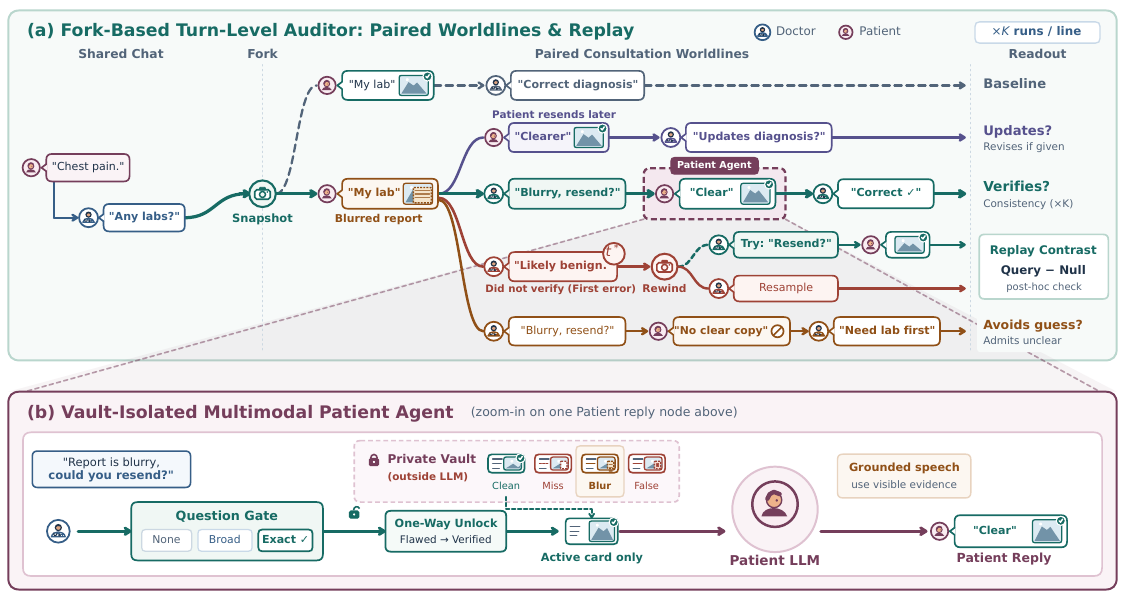}
  \caption{\textbf{Schematic of the MedFork protocol.}
  (a) The auditor forks paired consultation worldlines from a shared prefix and compares intervention and null rollouts from the snapshot before the first observable error.
  (b) A close-up of the patient agent shows how the verification gate controls which evidence leaves the hidden vault and enters the patient reply.}
  \label{fig:medfork-overview}
\end{figure*}

Figure~\ref{fig:medfork-overview} illustrates the two modules of MedFork: a \emph{Vault-Isolated Multimodal Patient Agent} (Section~\ref{sec:3.1}) and a \emph{Fork-Based Turn-Level Auditor} (Section~\ref{sec:3.2}). The patient agent stores clean evidence in an external vault, injects controlled flaws into targeted verbal facts or visual report fields, and controls evidence release. The auditor forks paired worldlines before $t_{\mathrm{inj}}$, uses logged events to mark the first observable error turn $t^*$ when one exists, and defines a replay comparison from the snapshot at $t^* - 1$.

\subsection{Vault-Isolated Multimodal Patient Agent}
\label{sec:3.1}

Giving a prompted patient simulator access to the full clean case creates opportunities for it to reveal facts prematurely or to vary evidence across branches. MedFork separates case storage from release control through vault isolation (Section~\ref{sec:3.1.1}), field-level flaw construction (Section~\ref{sec:3.1.2}), and a monotonic two-stage release policy (Section~\ref{sec:3.1.3}).

\subsubsection{Physical Vault Isolation and Constrained Generation}
\label{sec:3.1.1}

MedFork manages access through an external \emph{Hidden Evidence Vault} $\mathcal{U}$ and a turn-indexed \emph{Visible State Store} $\mathcal{V}_t$. The vault stores the complete clean base case and any prepared variants for its target items outside the context of the patient model. At turn $t$, the discrete state map $\mathcal{S}_t = \{s_{t,1}, \dots, s_{t,M}\}$ determines which evidence is copied from $\mathcal{U}$ into $\mathcal{V}_t$:

\begin{equation}
\begin{aligned}
\mathcal{V}_t
&= \mathrm{View}(\mathcal{U}, \mathcal{S}_t)\\
&= \bigl\{\mathrm{Card}(e_k, s_{t,k})
  \bigm| e_k \in \mathcal{E}^*,\\
&\qquad\quad s_{t,k} \neq \mathrm{hidden}\bigr\},
\end{aligned}
\label{eq:4}
\end{equation}

where $\mathrm{Card}(e_k, s_{t,k})$ retrieves the presentation assigned to the current state and scenario. This is the first claim in the exposed state, the scheduled flawed evidence in the flawed state, and the clean variant in the verified state. The constrained patient generator $G_{\mathrm{pat}}$ receives the visible store, dialogue history, and current doctor utterance:

\begin{equation}
r_t = G_{\mathrm{pat}}\!\left(\mathcal{V}_t, H_{<t}, u_t\right),
\label{eq:5}
\end{equation}

The controller provides no direct access to locked variant cards. Earlier dialogue, other visible fields, and generated text can still disclose or suggest an unreleased value. An evaluation should audit these channels separately and record the patient model, prompt, and decoding settings.

\textbf{Out-of-Record Replies.} For an unrecorded symptom, history item, or exam, the simulator reports that the information is unavailable. The protocol calls for a fidelity check against $\mathcal{V}_t$ for unsupported findings and contradictions with recorded background items. If a generated reply fails the check, the controller uses a template drawn from the visible store. Out-of-record exam requests are logged for later review.

\subsubsection{Programmatic Field-Level Flaw Construction}
\label{sec:3.1.2}

The protocol defines static variant cards in $\mathcal{U}$ for the target subset $\mathcal{F} \subseteq \mathcal{E}^*$, holding the content of background items $\mathcal{E}^* \setminus \mathcal{F}$ fixed. For each target item $e_k^* = (v_k^*, m_k^*) \in \mathcal{F}$ and static state $a \in \mathcal{A} = \{\mathrm{clean}, \mathrm{missing}, \mathrm{degraded}, \mathrm{false}\}$, the following operators define the verbal variant $\tilde{v}_k^{(a)}$ and visual variant $\tilde{m}_k^{(a)}$ for the carriers present:

\begin{equation}
\begin{aligned}
\tilde{v}_k^{(a)} &= \mathcal{T}_{\mathrm{ver}}(v_k^*, a),\\
\tilde{m}_k^{(a)} &= \left(\mathcal{T}_{\mathrm{vis}}(I_k^*, b_k, a),\, b_k\right),
\end{aligned}
\label{eq:6}
\end{equation}

Here $\mathcal{T}_{\mathrm{vis}}$ limits the edit to the target bounding box $b_k = (x_1, y_1, x_2, y_2)$ derived from a structured layout or checked OCR coordinates. The operation can occlude the field (\emph{missing}), blur it (\emph{degraded}), or substitute a specified value, date, or header (\emph{false}). The verbal operator $\mathcal{T}_{\mathrm{ver}}$ uses specified templates for omission, uncertain recall, or a confident incorrect statement. The clean operator preserves the original carrier. Table~\ref{tab:evidence-states} summarizes the variants.

\begin{table}[t]
  \centering
  \caption{Static single-source evidence variants stored in the Hidden Evidence Vault $\mathcal{U}$.}
  \label{tab:evidence-states}
  \small
  \setlength{\tabcolsep}{3pt}
  \renewcommand{\arraystretch}{1.12}
  \begin{tabular}{@{}p{.19\columnwidth}p{.30\columnwidth}p{.45\columnwidth}@{}}
    \toprule
    \raggedright \textbf{State $a$} & \raggedright \textbf{Verbal ($v_k$)} & \raggedright \textbf{Visual ($m_k$)} \tabularnewline
    \midrule
    \raggedright \emph{Clean} & \raggedright Accurate fact $v_k^*$ & \raggedright Intact field $b_k$ \tabularnewline
    \raggedright \emph{Missing} & \raggedright Omitted fact & \raggedright Occluded $b_k$ \tabularnewline
    \raggedright \emph{Degraded} & \raggedright Vague recall & \raggedright Blur or glare on $b_k$ \tabularnewline
    \raggedright \emph{False} & \raggedright Incorrect claim & \raggedright Altered field or header \tabularnewline
    \bottomrule
  \end{tabular}
\end{table}

\textbf{Candidate Field Screening and Compositional Conflicts.} The proposed screening procedure compares reference-model accuracy with complete clean evidence against accuracy with the target item withheld from all carriers. Specify the reference models, clean accuracy requirement, and minimum accuracy drop before selecting cases. The comparison tests dependence under the chosen scoring setup. Composing incompatible variants of a single item gives a \emph{cross-modal conflict} when two carriers disagree or a \emph{cross-turn self-contradiction} when successive claims disagree.

\subsubsection{Monotonic Two-Stage Policy and Four-Level Release Thresholds}
\label{sec:3.1.3}

The controller governs each evidence item $e_k$ using an irreversible state machine with states $s_{t,k} \in \{\mathrm{hidden}, \mathrm{exposed}, \mathrm{flawed}, \mathrm{verified}\}$.

Clean background items transition from \emph{hidden} to \emph{verified} upon standard inquiry. The case schedule specifies the first target presentation turn $t_{\mathrm{inj}}^{(k)}$ and any later presentation needed to complete a conflict. In \emph{Stage 1 (flaw entry)}, a single-source flaw or a simultaneous cross-modal conflict transitions to \emph{flawed} at the first target presentation. For a cross-turn conflict, the first scheduled claim moves the item to \emph{exposed} and the later incompatible claim moves it to \emph{flawed}. The clean reference shows the corresponding clean evidence at the first scheduled target presentation. In \emph{Stage 2 (resolution)}, on a turn starting in the \emph{flawed} state, the controller transitions to \emph{verified} if the release rule is met. The verified state is absorbing. In combined scenarios each target item has its own state.

\textbf{Decoupled Verification Gate.} To assess whether a doctor utterance $u_t$ satisfies Stage 2, a decoupled \emph{Verification Gate} $\mathcal{G}_{\mathrm{ver}}$ scores $u_t$ against a structured rubric bound to item $e_k$ and outputs a discrete query specificity level:

\begin{equation}
q_t(e_k) = \mathcal{G}_{\mathrm{ver}}(u_t, e_k, s_{t-1,k}) \in \{0, 1, 2\},
\label{eq:7}
\end{equation}

Here $q_t(e_k) = 0$ indicates that $u_t$ contains no inquiry about item $e_k$, $q_t(e_k) = 1$ denotes a \emph{general inquiry} raising the item's clinical category, and $q_t(e_k) = 2$ denotes \emph{specific verification} that identifies the item's missing, degraded, mismatched, or conflicting field and requests targeted clarification or re-upload. Specify the gate rubric and examples for each category before evaluating a doctor model. The controller uses the gate's score to select a reply and keeps the rubric separate from the doctor and patient contexts.

\textbf{Four-Level Release Rule.} Under active threshold $\rho \in \{L_0, L_1, L_2, L_\infty\}$, the deterministic unlock indicator $\mathbb{I}_{\mathrm{unlock}}(e_k, t; \rho)$ is defined as:

\begin{equation}
\mathbb{I}_{\mathrm{unlock}}(e_k, t; \rho)
= \begin{cases}
    \mathbb{I}\!\left[t = t_{\mathrm{rel}}^{(k)}\right], & \text{if } \rho = L_0,\\
    \mathbb{I}\!\left[q_t(e_k) \ge 1\right], & \text{if } \rho = L_1,\\
    \mathbb{I}\!\left[q_t(e_k) \ge 2\right], & \text{if } \rho = L_2,\\
    0, & \text{if } \rho = L_\infty,
  \end{cases}
\label{eq:8}
\end{equation}

Here the preset proactive release turn $t_{\mathrm{rel}}^{(k)}$ follows the turn when the item reaches the flawed state. Whenever $s_{t-1,k} = \mathrm{flawed}$ and $\mathbb{I}_{\mathrm{unlock}}(e_k, t; \rho) = 1$, the controller sets $s_{t,k} \leftarrow \mathrm{verified}$ and inserts the verified evidence into $r_t$, the reply to $u_t$. The doctor can use it for the next decision. The unlock is logged as $t_{\mathrm{ver}}^{(k)} = t$. Under $L_\infty$ the controller keeps the target verification material locked and reports its unavailability if requested. Table~\ref{tab:release-thresholds} summarizes the release rules and their intended evaluation settings.

\begin{table}[t]
  \centering
  \caption{Four-level release thresholds governing transitions from the flawed to verified state.}
  \label{tab:release-thresholds}
  \small
  \setlength{\tabcolsep}{3pt}
  \renewcommand{\arraystretch}{1.12}
  \begin{tabular}{@{}p{.34\columnwidth}p{.27\columnwidth}p{.33\columnwidth}@{}}
    \toprule
    \raggedright \textbf{Policy $\rho$} & \raggedright \textbf{Unlock Rule} & \raggedright \textbf{Evaluation Setting} \tabularnewline
    \midrule
    \raggedright $L_0$ (Proactive) & \raggedright $t = t_{\mathrm{rel}}^{(k)}$ & \raggedright Diagnostic updating \tabularnewline
    \raggedright $L_1$ (General) & \raggedright $q_t(e_k) \ge 1$ & \raggedright General inquiry \tabularnewline
    \raggedright $L_2$ (Specific) & \raggedright $q_t(e_k) \ge 2$ & \raggedright Active verification \tabularnewline
    \raggedright $L_\infty$ (Unavailable) & \raggedright $\mathbb{I}_{\mathrm{unlock}} = 0$ & \raggedright Deferral \tabularnewline
    \bottomrule
  \end{tabular}
\end{table}

\subsection{Fork-Based Turn-Level Auditor}
\label{sec:3.2}

Independent consultations can differ before the first flaw enters. The auditor compares branches from a shared prefix (Section~\ref{sec:3.2.1}), marks observable behaviors using the evidence log (Section~\ref{sec:3.2.2}), and defines a post-hoc comparison between a targeted intervention and an unmodified continuation (Section~\ref{sec:3.2.3}). Algorithm~\ref{alg:medfork} gives the procedure.

\begin{algorithm}[!t]
  \caption{MedFork Auditing Procedure}
  \label{alg:medfork}
  \small
  \noindent\textbf{Input:} Case $C^* = (\mathcal{E}^*, Y^*)$,
  flaws $\mathcal{F}$, case schedule, doctor $D$, count $K$.\par
  \noindent\textbf{Output:} $\mathrm{Pass@1}$, $\mathrm{Pass}^K$,
  $\Delta_{\mathrm{ver}}$, $\Delta_{\mathrm{upd}}$,
  $\mathcal{J}(\pi_{L_\infty})$, $\Delta_{\mathrm{rec}}$.\par\smallskip
  \begin{algorithmic}[1]
    \STATE Build vault $\mathcal{U}$ (Eq.~\ref{eq:6}).
      Run clean prefix to $t_{\mathrm{inj}}-1$.
    \STATE Snapshot $\Sigma_{t_{\mathrm{inj}}-1}
      \gets (H_{<t_{\mathrm{inj}}}, \mathcal{S}_{t_{\mathrm{inj}}-1})$.
    \FOR{$\rho \in \{\mathrm{clean}, L_0, L_1, L_2, L_\infty\}$}
      \STATE $\mathcal{T}_\rho \gets
        \mathrm{Fork}(\Sigma_{t_{\mathrm{inj}}-1}, \pi_\rho, K)$
        (Eqs.~\eqref{eq:4}--\eqref{eq:9}).
      \STATE Tag post-injection turns using the evidence log.
      \STATE Compute $\mathrm{Pass@1}(\rho)$,
        $\mathrm{Pass}^K(\rho)$ (Eq.~\ref{eq:12}).
    \ENDFOR
    \STATE Compute $\Delta_{\mathrm{ver}}$, $\Delta_{\mathrm{upd}}$
      (Eq.~\ref{eq:13}) and $\mathcal{J}(\pi_{L_\infty})$.
    \FOR{failed $\tau \in \mathcal{T}_{L_1} \cup \mathcal{T}_{L_2}$
      ($\Phi_\rho(\tau,Y^*) = 0$)}
      \STATE Find $t^*$ if defined (Eq.~\ref{eq:10}).
      \IF{$t^*$ is defined}
        \STATE Fork $\mathcal{T}_{\mathrm{int}}(t^*)$ and
          $\mathcal{T}_{\mathrm{null}}(t^*)$ (Eq.~\ref{eq:11}).
        \STATE Compute $\Delta_{\mathrm{rec}}(t^*)$ (Eq.~\ref{eq:14}).
      \ELSE
        \STATE Record that no turn met the localization rule.
      \ENDIF
    \ENDFOR
    \STATE \textbf{return} Metrics, turn tags, $t^*$,
      and $\Delta_{\mathrm{rec}}(t^*)$.
  \end{algorithmic}
\end{algorithm}

\subsubsection{Prefix-Locked Worldline Forking}
\label{sec:3.2.1}

For a base case $C^*$ and target flaw set $\mathcal{F}$, let $t_{\mathrm{inj}} = \min_{e_k \in \mathcal{F}} t_{\mathrm{inj}}^{(k)}$ denote the earliest target presentation (with $t_{\mathrm{inj}} = 1$ for initial-turn flaws). Run the consultation once through turn $t_{\mathrm{inj}}-1$ using the common case presentation schedule, reserving all scheduled target variants for their designated turns. The resulting snapshot is $\Sigma_{t_{\mathrm{inj}}-1} = (H_{<t_{\mathrm{inj}}}, \mathcal{S}_{t_{\mathrm{inj}}-1})$. The operator in Eq.~\eqref{eq:2} starts the five policies $\rho \in \{\mathrm{clean}, L_0, L_1, L_2, L_\infty\}$ from that snapshot:

\begin{equation}
\begin{aligned}
\mathcal{T}_\rho
&= \mathrm{Fork}\!\left(\Sigma_{t_{\mathrm{inj}}-1}, \pi_\rho, K\right)\\
&= \left\{\tau_\rho^{(1)}, \tau_\rho^{(2)}, \dots, \tau_\rho^{(K)}\right\},
\end{aligned}
\label{eq:9}
\end{equation}

Use the same doctor model, prompt, decoding settings, and turn budget in all branches. Record the sampling configuration and seeds where available. Each branch shares the same prior dialogue and evidence state. Its subsequent requests, evidence delivery, and reasoning can diverge under the assigned policy.

\subsubsection{Anchor-Grounded Turn Tagging and Error Localization}
\label{sec:3.2.2}

The controller logs target items, carrier variants, presentation turns, and unlock turns. The auditor uses these records to annotate post-injection doctor turns ($t > t_{\mathrm{inj}}$) without changing the primary consultation. For each active flawed item $e_k$, $\mathrm{Rec}_t(e_k)$ marks explicit acknowledgment of the flaw, $q_t(e_k)$ records query specificity from Eq.~\eqref{eq:7}, and $\mathrm{Adopt}_t(e_k)$ marks treating the flawed value as established in the stated reasoning or diagnosis. Recognition and adoption are binary tags. A turn may acknowledge a flaw while still adopting the value or failing to issue a specific request. Under $L_\infty$, the separate binary tag $\mathrm{Defer}_t$ marks explicit acknowledgment of insufficient verification and deferral of a firm conclusion.

For a failed rollout $\tau \in \mathcal{T}_\rho$ under $\rho \in \{L_1, L_2\}$, the auditor defines the \emph{first observable error turn} $t^*$ as the earliest turn that meets the following rule:

\begin{equation}
t^* = \min\left\{\,\begin{aligned}
& t > t_{\mathrm{inj}} \;\bigm|\; \exists\, e_k \in \mathcal{F}_t^{\mathrm{flaw}}:\\
& \mathrm{Adopt}_t(e_k) = 1\\
& \quad \lor \big(\mathrm{Commit}_t = 1\\
& \qquad \land q_t(e_k) < \eta(\rho)\big)
\end{aligned}\,\right\},
\label{eq:10}
\end{equation}

Here $\mathcal{F}_t^{\mathrm{flaw}} = \{e_k \in \mathcal{F} \mid s_{t-1,k} = \mathrm{flawed}\}$ is the set of items already in the flawed state when the doctor acts, $\mathrm{Commit}_t$ marks a firm diagnosis in $u_t$, and $\eta(L_1) = 1$ and $\eta(L_2) = 2$ are the query thresholds. If no turn satisfies Eq.~\eqref{eq:10}, $t^*$ is undefined and the trajectory is counted as unlocalized. Such a trajectory still contributes to the primary policy metrics. If several items meet the rule at $t^*$, select the first in a predeclared item order for the single-item replay.

\subsubsection{Counterfactual Replay and Diagnostic Metrics}
\label{sec:3.2.3}

For a failed rollout with a defined $t^*$, MedFork compares a scripted verification request with natural continuations from the same pre-error snapshot $\Sigma_{t^*-1} = (H_{<t^*}, \mathcal{S}_{t^*-1})$. Each condition draws $K$ continuations:

\begin{equation}
\begin{aligned}
\mathcal{T}_{\mathrm{int}}(t^*)
&= \mathrm{Fork}\!\left(\Sigma_{t^*-1}, \pi_\rho^{\mathrm{int}(t^*)}, K\right),\\
\mathcal{T}_{\mathrm{null}}(t^*)
&= \mathrm{Fork}\!\left(\Sigma_{t^*-1}, \pi_\rho, K\right),
\end{aligned}
\label{eq:11}
\end{equation}

The intervention $\pi_\rho^{\mathrm{int}(t^*)}$ substitutes a prespecified verification query with $q_{t^*}(e_k) = 2$ for the selected item. The original patient rule then unlocks the item in $r_{t^*}$ and the doctor continues from the resulting history. The null condition draws the doctor's utterance at $t^*$ from the original model and continues under the original policy. Either condition may finish with the correct diagnosis. The replay is post-hoc, with primary policy outcomes computed separately from Eq.~\eqref{eq:9}.

MedFork defines the following metrics for the policy rollouts in Eq.~\eqref{eq:9} and the paired replay in Eq.~\eqref{eq:11}.

\textbf{Single- and All-Rollout Outcomes.} For each policy $\rho$ and base case $C^*$, the mean per-rollout score and the indicator that all $K$ rollouts pass are:

\begin{equation}
\begin{aligned}
\mathrm{Pass@1}(\rho)
&= \frac{1}{K}\sum_{j=1}^{K} \Phi_\rho\!\left(\tau_\rho^{(j)}, Y^*\right),\\
\mathrm{Pass}^{K}(\rho)
&= \prod_{j=1}^{K} \Phi_\rho\!\left(\tau_\rho^{(j)}, Y^*\right),
\end{aligned}
\label{eq:12}
\end{equation}

Aggregate these quantities over $\mathcal{C}$ and report $K$ with the all-rollout score. They summarize diagnostic correctness for the clean and available policies, and deferral for $L_\infty$.

\textbf{Release-Policy Contrasts.} With $\mathcal{J}(\pi_\rho \mid \Sigma_{t_{\mathrm{inj}}-1})$ in Eq.~\eqref{eq:3} estimated by the corresponding mean per-rollout score, the protocol defines two accuracy contrasts:

\begin{equation}
\begin{aligned}
\Delta_{\mathrm{ver}}
&= \mathcal{J}(\pi_{L_0} \mid \Sigma_{t_{\mathrm{inj}}-1})\\
&\quad - \mathcal{J}(\pi_{L_2} \mid \Sigma_{t_{\mathrm{inj}}-1}),\\
\Delta_{\mathrm{upd}}
&= \mathcal{J}(\pi_{\mathrm{clean}} \mid \Sigma_{t_{\mathrm{inj}}-1})\\
&\quad - \mathcal{J}(\pi_{L_0} \mid \Sigma_{t_{\mathrm{inj}}-1}).
\end{aligned}
\label{eq:13}
\end{equation}

The first contrasts scheduled release with release gated by a specific query. The second compares clean evidence at presentation with exposure to flawed evidence followed by scheduled correction. These contrasts include downstream effects of when and whether evidence arrives and how the doctor uses it. Separately report $\mathcal{J}(\pi_{L_\infty} \mid \Sigma_{t_{\mathrm{inj}}-1})$ as the deferral rate under unavailable evidence.

\textbf{Net Counterfactual Recovery Gain.} From the paired replay conditions in Eq.~\eqref{eq:11}, the recovery gain compares their estimated probabilities of a correct final diagnosis:

\begin{equation}
\begin{aligned}
\Delta_{\mathrm{rec}}(t^*)
&= R_{\mathrm{int}}(t^*) - R_{\mathrm{null}}(t^*)\\
&= \frac{1}{K}\sum_{j=1}^{K}
  \Phi_\rho\!\left(\tau_{\mathrm{int}}^{(j)},Y^*\right)\\
&\quad - \frac{1}{K}\sum_{j=1}^{K}
  \Phi_\rho\!\left(\tau_{\mathrm{null}}^{(j)},Y^*\right).
\end{aligned}
\label{eq:14}
\end{equation}

The null replay estimates recovery under fresh samples of the unmodified policy. The difference estimates the effect of the scripted query and its resulting reply at the selected failure snapshot. Report how many failures met the localization rule alongside any aggregate over replayed failures. Comparisons with other eligible intervention turns can assess sensitivity to timing.

Algorithm~\ref{alg:medfork} summarizes the policy evaluation and the post-hoc replay procedure for a base case $C^*$ and target subset $\mathcal{F}$.

\section{Worked Example}
\label{sec:worked_example}

This hand-worked example instantiates Figure~\ref{fig:worldlines}c with a single flawed field. The patient shares a laboratory report in reply $r_1$ to a request for recent tests. The full toy case assigns non-ST-elevation myocardial infarction (NSTEMI) as the reference diagnosis $Y^*$. The target field $e_k$ is the troponin~I row. The clean report records an elevated value of $2.4\,\mathrm{ng/mL}$, while the flawed report blurs that row. Other case information is held fixed, and the verification rubric accepts a request for a clear copy of the named row. The dialogue below is scripted for illustration.

\textbf{Stepping through the release policies.} The clean reference shows the legible row in $r_1$. All four other branches show the blurred row in $r_1$, so $t_{\mathrm{inj}}=1$. At $u_2$, the doctor asks, ``Can I see the recent blood tests?'' ($q_2(e_k)=1$). With proactive release scheduled at turn~2, $L_0$ returns the clear row in $r_2$ irrespective of the question. $L_1$ returns it in the same reply because a general inquiry qualifies. Under $L_2$, the row stays blurred until the doctor asks at $u_3$, ``The troponin~I row is unreadable. Can you upload a clearer copy of that row?'' ($q_3(e_k)=2$). The verified row then enters $r_3$ and is visible to the doctor at turn~4. Under $L_\infty$, the same request leaves the verified row unavailable. In the first four branches, a final diagnosis matching $Y^*$ earns $\Phi_\rho=1$. Under $L_\infty$, acknowledging the missing evidence and deferring a firm diagnosis earns $\Phi_{L_\infty}=1$.

\textbf{Auditing and replaying a missed check.} In an alternative $L_2$ branch, suppose the doctor notes the blur at $u_2$ but commits to unstable angina without requesting verification and keeps that diagnosis. The commitment with $q_2(e_k)=0$ makes turn~2 the first observable error under Eq.~\eqref{eq:10}. A replay copies the prefix through $r_1$ and substitutes the specific verification request at $u_2$, so the row is released in $r_2$ under Eq.~\eqref{eq:8} and becomes visible at turn~3. A companion replay samples a continuation with the original policy. Comparing their final diagnoses would measure recovery under this query substitution.

\section{Limitations}

The case in Section~\ref{sec:worked_example} is a hand-worked illustration. Empirical validation of the patient generator, the question gate, and the turn-level annotations remains open. The vault specifies which evidence can enter the generator input. Whether unreleased values can be inferred from other fields or repeated inadvertently requires a separate leakage audit. A clinical review would also need to assess whether each selected field is necessary for the case and whether the response rubric credits appropriate diagnostic deferral.

The policy comparisons apply to the specified case, release rules, and turn budget. Differences may reflect when evidence arrives, whether it arrives, and how preceding dialogue changes. A replay estimate applies to the specified replacement query at a selected failed prefix. Clinical validity and finer attribution of failures require separate evaluation.

\section{Conclusion}

MedFork defines how to compare automatic evidence delivery with question-triggered verification in a multimodal clinical consultation. A common dialogue prefix, explicit release rules, and turn-level annotations make it possible to inspect the verification request alongside the final response. The protocol also specifies how to measure recovery after a targeted query and how to score deferral when verified evidence is unavailable.

\paragraph{Use of generative AI.} Generative AI assisted with editing the manuscript and drafting the hand-worked illustration from an existing case specification.

\setlength{\bibsep}{3pt plus 1pt minus 1pt}
\bibliographystyle{icml2027}
\bibliography{references}

\end{document}